# Understanding colors of Dufaycolor:
## Can we recover them using historical colorimetric and spectral data?

Jan Hubička[1], Linda Kimrová[2], Melichar Konečný[2]

[1]Department of Applied Mathematics, Charles University; Šechtl & Voseček Museum of Photography; SUSE ČR s.r.o.

[2]Charles University

Contact: Jan Hubička, hubicka@kam.mff.cuni.cz

**Abstract**

Dufaycolor, an additive color photography process produced from 1935 to the late 1950s, represents one of the most advanced iterations of this technique. This paper presents ongoing research and development of an open-source Color-Screen tool designed to reconstruct the original colors of additive color photographs. We discuss the incorporation of historical measurements of dyes used in the production of the color-screen filter (réseau) to achieve accurate color recovery.

**Introduction**

*Additive color processes* of early color photography employed a variety of techniques to filter light into its primary components – red, green, and blue – and then recombine them to create a full-color image. A common approach, proposed by Louis Ducos du Hauron in his seminal 1869 work "*Les Couleurs en Photographie, solution du problème*," involved the use of a fine screen or mosaic of color filters (referred to here as a *color-screen*) placed in front of a light-sensitive monochromatic emulsion. The first commercially available process based on this technique was introduced by John Joly in 1895. Number of improved processes followed (Pénichon, 2013), including the most widely known one, Autochrome Lumiére introduced in 1907 (Lavédrine and Gandolfo, 2013).

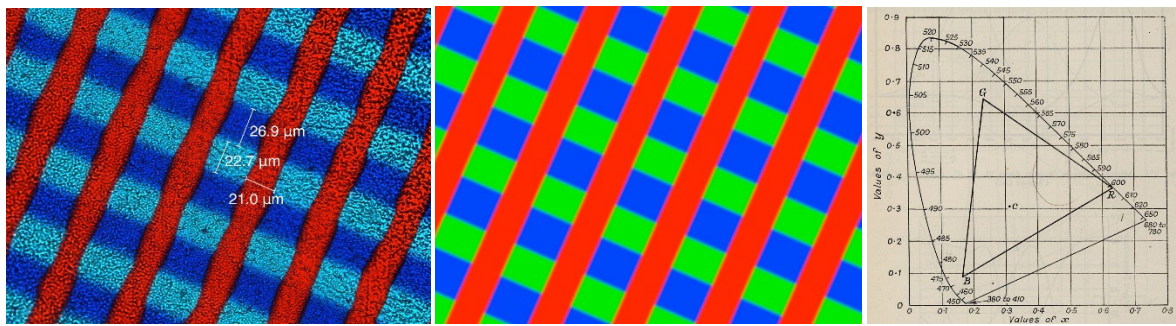

Fig. 1 – Dufaycolor réseau: (a) microscopic image[1], (b) digital simulation of original colors, (c) primaries on the chromaticity diagram (Cornwell-Clyne, 1951) p150.

Dufaycolor, introduced in 1935, was a relatively late example of an additive color process, but also among the most refined and best documented. The Dufaycolor film consisted of a transparent film base coated with a color-screen filter called a *réseau*. This réseau was a mosaic comprised of tiny, alternating green and blue squares, interleaved with red lines (Fig. 1). (We call those miniature filters *elements*.) A layer of protective varnish separated the réseau from a panchromatic photographic emulsion. The film was loaded into the camera with the base facing the lens, the reverse of conventional practice. During the exposure, light passed through the film base and then through the colored réseau elements before reaching the panchromatic emulsion. Each element of the réseau acted as a tiny filter, allowing only its specific color of light to pass through. This resulted in a record of the color information behind each element being captured by the underlying panchromatic emulsion. This principle is conceptually similar to that used in modern digital cameras, where a Bayer filter, a modern color-screen, is placed over a monochromatic CMOS or CCD sensor. The réseau served a dual role in the Dufaycolor process. Firstly, it encoded the color information during the initial exposure, as described above. Secondly, after the photographic emulsion was developed into a high-contrast monochromatic transparency (an *image layer*), the réseau in combination with the image layer visually recreated the recorded colors.

---

[1] The Emulsion has been removed. The visible structures are not silver grain but the structure of the filter layers. From the Timeline of Historical Film Colors, image by David Pfluger, ERC Advanced Grant FilmColors. Imaging was performed with support of the Center for Microscopy and Image Analysis, University of Zurich.

The dyes used in the réseau were not particularly stable and lightfast, causing virtually all surviving Dufaycolor images to appear faded today. However, the silver-based image layer is significantly more stable. This allows for the possibility of digital restoration through infrared scanning: Capturing the color scene in image layer gave an analog variant of a RAW file from a modern digital camera. Once digitized (réseau superposing the image layer is transparent in the infrared light), identifying the individual color-screen elements within the scan becomes a critical yet challenging step due to factors like deformations caused by shrinking of the film base, and geometric distortions introduced during scanning. Nevertheless, with these challenges addressed, modern image processing techniques, such as demosaicing algorithms and color-space conversions, can be applied to reconstruct a full-color image from the recovered data with greater level of precision then possible from faded subtractive films (Barker *et al.*, 2022; Hubička *et al.*, 2023).

We are developing an open-source tool, Color-Screen[2], specifically designed for the analysis and color reconstruction of early color photographs captured using color-screen methods. This tool has successfully yielded well looking color reconstructions from a variety of processes, including *Paget*, *Dioptichrome*, *Finlay* color plates as well as Dufaycolor film. This paper focuses on the challenges encountered in achieving faithful color reconstruction, aiming to reproduce the colors as they appeared when the photographs were new, before the degradation of dyes within the color-screen filters. This task involves understanding numerous technical details. While the method is applicable to other color-screen processes, this paper will primarily examine Dufaycolor due to its comparatively extensive documentation, with plans to extend our research to other processes in the future. Because all Dufaycolor photographs are significantly faded today, we dive into historical records on Dufaycolor film to find data for the digital simulations.

**Manufacturing Dufaycolor's réseau**
Examining Dufaycolor film in a microscope reveals the surprisingly precise printing of the réseau, with individual color elements arranged without significant overlaps or gaps that could compromise color fidelity. While some variation in the uniformity of these elements exists, their high density effectively averages out these minor inconsistencies. Multiple types of réseau screens were produced across different production lines. The density of the lines ranged approximately from 19 to 25 lines per millimeter (480 to 635 lines per inch). This more than doubles the line density of modern high-quality halftone prints, which typically use around 200 lines per inch. The high density of the réseau's color elements is attributable to the process's original development for motion picture film, where such resolution was necessary for satisfactory image quality upon projection.

(Cornwell-Clyne, 1951), in pages 285–290, gives what we believe to be the most comprehensive description of the réseau's manufacturing process, including photographs of the printing machinery. We refer the interested reader to this publication for a detailed account and (Friedman, 1947) for a survey of many associated patents. In brief, the process involved first coating the acetate film base with a layer of blue-dyed collodion. Subsequently, the color pattern of the réseau was created in a two-step process using a specialized machine similar to a halftone press. This machine was capable of printing fine lines of greasy ink, transferred from an engraved cylindrical roller.

---

[2] https://github.com/janhubicka/Color-Screen/wiki

The *first step* of the process involved printing a series of lines at an angle of approximately 23° to the film's edge. The areas of the blue-dyed collodion not covered by these lines were then bleached and dyed green. This created a pattern of alternating green and blue stripes. In the *second step*, a different engraved cylinder was used, featuring lines oriented roughly, but not precisely, orthogonal to the first set, and with a similar, but again not exact, line density. After this second printing, the remaining exposed blue- and green-dyed collodion was bleached and dyed red, resulting in the final, characteristic réseau pattern. This process is also apparent in microscope, since the blue lines appears closest to the film base, green lines in between and red lines most distant.[3] Final steps of production line involved application of the protection layer and finally coating by the photographic emulsion.

Research on Autochrome plates (Lavédrine, 1992) has revealed significant variations in the colors of starch grains producing Autochrome's color-screen and their different percentages across different production batches.[4] These inconsistencies yield doubts on the possibility of achieving reliable color reconstruction for Autochrome images. Dufaycolor, however, was designed for both still and motion picture film, with the latter imposing far stricter demands for reproducibility. Consequently, it is reasonable to expect that Dufaycolor réseau production exhibited greater uniformity. This hypothesis is further supported by the comparatively detailed quality control procedures (Cornwell-Clyne, 1951):

Following the first printing step, the film underwent quality control in a view-box illuminated by CIE Standard Illuminant B.[5] This inspection involved assessing the straightness, evenness, and number of breaks in the printed lines. Additionally, the degree of ink penetration and the frequency of printing anomalies were evaluated. After the second printing stage, the color accuracy of the réseau was measured using a colorimeter and compared against a set of tricolor filters. If the réseau's color deviated beyond a (regrettably unspecified) tolerance from a target chromaticity (also unfortunately unknown), the film was back-coated with a subtractive filter to adjust its color properties.

|  | x | y | Y | Dom. Wavelength |
|---|---|---|---|---|
| **Red element** | 0.633 | 0.365 | 17.7 | 601.7 |
| **Green element** | 0.233 | 0.647 | 43 | 549.6 |
| **Blue element** | 0.14 | 0.089 | 3.7 | 466.0 |

Tab. 1 - Colorimetric specification of réseau elements, (Cornwell-Clyne, 1951) page 290.

The colorimetric specifications are given in Tab. 1. We find it remarkable that the 1931 CIE standard has been already applied in production of Dufaycolor. (Bonamico and Baker, 1933) also mentions: "*It is scarcely necessary to say that every new batch of dyes used has to be tested with the spectrophotometer, so that its concentration may be modified in accordance with requirements*."

**Digital simulation of the réseau and problem in the published data**
To simulate the réseau filter digitally, we combined data in Tab. 1 with measurements of the red, green, and blue elements' dimensions in Fig. 1 (a). While the width of the blue and green squares is not given, it can be estimated to be approx. 28.9 μm. Based on this we determined that red filter covers 42.1 % of area, green filter 26.5 % and blue 31.4 %. Simulated réseau is shown in Fig. 1 (b).

---

[3] Réseau printed with blue or green color last exists as well, those are used by earlier Spicer-Dufay motion film.
[4] Lavédrine, private communication.
[5] Illuminant B is an obsolete standard with tungsten light corrected to represent direct sunlight at intermediate and northern latitudes, with a correlated color temperature of approximately 4874 K. Later superseded by Illuminant D.

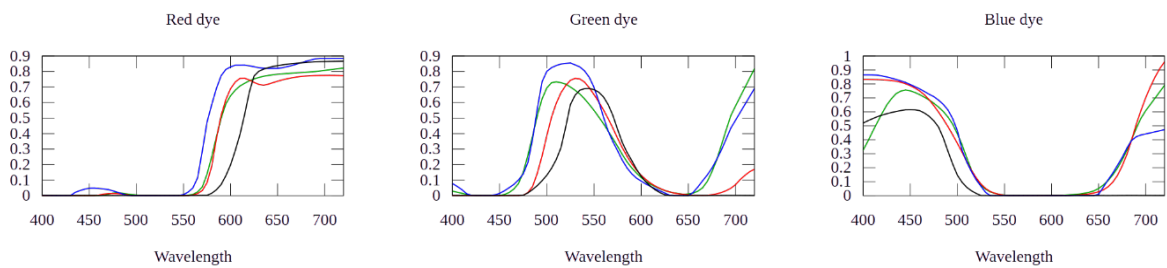

Fig. 2 – Transmission curves of primaries réseau from different publications. Green: (Harrison and Horner, 1939) p325, red: (Cornwell-Clyne, 1951) p149, blue: (Collins and Giles, 1952) p435 and black: (Neblette, 1952) p446.

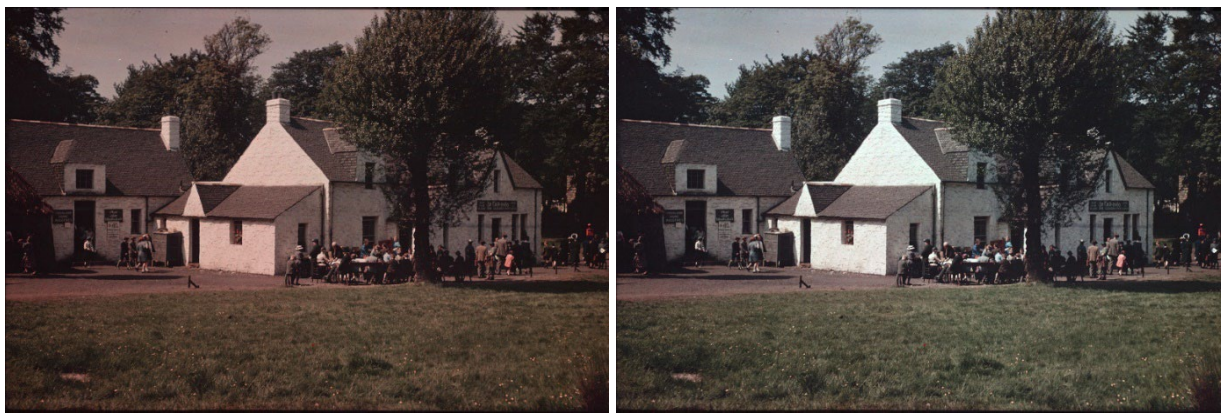

Fig. 3 – Sample Dufaycolor 6×9 cm approx. 1937, Clachan Inn, Scottland: (a) color calibrated scan, (b) with color of réseau balanced to neutral using levels function in GIMP. Probably taken by J.B.S. Thubron and/or his wife Diana.

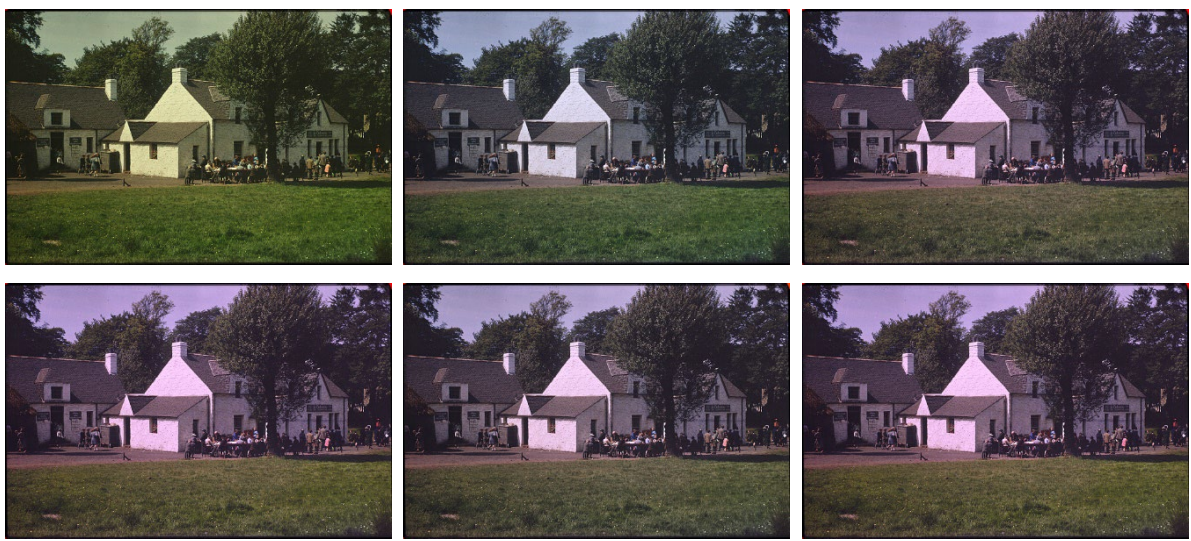

Fig. 4 – Color reconstructions using réseau primaries using xyY specification of primaries: (a) Tab. 1, (b) Tab. 2; and using spectra from Fig. 1: Harrison and Horner (c), Cornwell-Clyne (d), Collins-Giles (e), and Neblette (f).

Simulating color of our sample Dufaycolor scans using this réseau screen led to poor results, see Fig. 4 (a), with a relatively simple explanation. Additively combining the colors of the réseau elements in the proportions derived from their measured dimensions results in a greenish color. While Tab. 1 does not specify the illuminant used for the color measurements, it was likely either Illuminant B or C, both intended to simulate daylight conditions. Furthermore, the original color specification of the combined réseau is unknown but close to neutral grey. If réseau had a saturated overall color it would show in all over-exposed areas of photographs which would be undesirable (Harrison and Horner, 1937; Neblette, 1952). This can be verified. Infrared imaging can be used to measure the visual density of the image layer in areas of a photograph we believe to be originally neutral grey and in these areas the densities are approximately the same whatever the colour of the reseau patch.

To verify Tab. 1 we recomputed dominant wavelengths (those are computed using chromaticity coordinates of the color and whitepoint). No whitepoint could reconcile the published values in Tab. 1. Examining other tables in the same source (Cornwell-Clyne, 1951) and tracing them back to their original sources revealed a relatively high frequency of typographical errors. This unfortunate fact can be attributed to the huge quantity of scientific data presented in the book.

Unfortunately, the original source of Tab. 1 could not be located. Luckily, page 150 of (Cornwell-Clyne, 1951), contains chromaticity diagram of the réseau elements, see Fig. 1 (c). Red and green primaries measured from this diagram align with the values in Tab. 1 (up to 0.005). The blue primary, however, suggests a missing digit in Tab. 1; specifically, that the x-coordinate of 0.14 should be 0.164 so all quantities are with 3 digit precision. Furthermore, to achieve a more neutral overall color when combining the réseau elements, we found it necessary to adjust the intensity of the blue primary from 3.7 % to 8.7 %. These corrections are summarized in Tab. 2 and corresponding rendering in Fig. 4 (b). Still, we failed to match the dominant wavelength of green element as indicated.

|  | x | y | Y | Dom. Wavelength |
|---|---|---|---|---|
| **Red element** | 0.633 | 0.365 | 17.7 | 601.7 |
| **Green element** | 0.233 | 0.647 | 43 | ~~549.6~~ **534.8** |
| **Blue element** | ~~0.14~~ **0.164** | 0.089 | ~~3.7~~ **8.7** | 466.0 |

Tab. 2 - Attempted correction to the colorimetric specification of réseau elements. Dominant wavelengths are computed relative to Illuminant E (0.333 x, 0.333 y) which works well for red and blue.

**Dyes used to print réseau**

Motivated by the necessity to verify Tab. 2 we searched for information about dyes used in réseau. Fascinating insight into this area, given by (Bonamico and Baker, 1933), explains, among others, why the dyes used in réseau seems less stable than ones in Autochrome:

*"The dye problem has been a difficult one. Owning to the chemical nature of the bleaching baths, it is necessary to use basic dyes throughout, and certain dyes, e.g. Methylene blue, will migrate through the protective layer to the photographic emulsion and cause innumerable small insensitive spots, which, on reversal, appear as black spots. A large amount of work has been carried out on the basis of trial and error because, although we can make up dye-baths to conform to any theoretical specification, we are obliged to employ dyes which will not migrate to emulsions. Moreover, they must be sufficiently soluble to obtain the very high concentrations necessary for efficient colouring to take place at the very rapid rate at which film passes through the machine. The dyes used must withstand also the repeated process to which the matrix is subjected. The thickness of cinematograph film is 5/1000 in., and the coating of green collodion has to be put on top of it[6], as well as the protective layer of gelatin, followed by a second protective layer of varnish, and finally the emulsion. It will be seen, therefore, that, unless these layers represent the absolute minimum of thickness, the film will be so unwieldy that it would not pass through the slender gate of the cine-camera and projector. Nevertheless, in this extremely thin film of collodion it is necessary to concentrate the requisite amount of all three dyes in order to give the required absorption within close physical limits. One way in which the colour chemist can help to improve the brilliancy of the pictures is to provide dyes, which are of the highest possible efficiency for the purpose under discussion, i.e. dyes which will give the necessary transmission of colour, but will not absorb a large proportion of the incident light. The fugitive character of the dyes is not of great importance, since the exposure of an individual picture frame in the projector only occupies 1/40th sec., and the life of average film is about 200 projections. The actual exposure of the colours, therefore, to the light of the arc lamp is of the order of 5 sec., so that the property of the permanence can be ignored."*

---

[6] Published in 1933 this text refers to the pre-production variant of earlier Spicer-Dufay process or Dufaycolor. Spicer-Dufay réseau is printed with blue color last and Dufaycolor with red last. Also protection layer is likely synthetic resin.

We located four independent sources for spectral transmission curves of Dufaycolor primaries (summarized in Fig. 2). However, definitive information on the exact dye formulations proved elusive. While US patent 1,806,361 (filed 1931) describes a printing process using specific dye mixtures—*blue* dye (100 parts of alcohol, 4 parts of *malachite green*, 6.7 parts of *auramine)*, *red* dye (*basic red N Extra*, Kühlmann), *blue*/violet dye (80 parts 4% *crystal violet* 4% in alcohol, 20 parts 8% *malachite gre*en in alcohol)—these formulations, also cited in (Collins and Giles, 1952) and (Friedman, 1947), are examples rather than precise recipes, possibly altered to protect proprietary information. These described mixtures do not appear to correspond with the curves shown in Fig. 2.

**Experimental results**

Simulating the Dufaycolor réseau using spectral data requires selecting a backlight. (*The Dufaycolor Manual.* first edition, 1938) specifies that the réseau elements are balanced for viewing under normal daylight. Consequently, we used the D65 illuminant in our simulation. (Color-Screen allows selection of other color temperatures and applies a Bradford correction to the whitepoint of the output color space.) All additive transparencies appear dark without adequate backlighting. It was common practice to mount these transparencies in wide black mats or use specialized viewers that blocked all light not passing through the transparency. The results of our simulation are shown in Fig. 4 (c)–(f).

To evaluate significance of the data about réseau primaries we used sample Dufaycolor transparencies digitized using Nikon Coolscan 9000ED at 4000PPI in RGB and infrared channel. Color-Screen tool, using the RAW data obtained, performed later processing. First it analyzed the geometry of the scan (used RGB scan to identify individual color elements of réseau). Next it simulated the scanner scanning the réseau to determine dot spread functions at multiple spots of the scan. (This is necessary to compensate for variation of sharpness across the scan caused by film deformation). Next the infrared channel was used to determine intensities of each color element and image was demosaiced by bicubic interpolation. Using the information about dot spread function in multiple points of the scan the saturation loss is estimated and compensated for obtaining an RGB image in the color profile of Dufaycolor film. Finally, given primary colors are applied, and image is converted to XYZ. Overall brightness (exposure) was adjusted according to the simulated réseau has Y of 1.

|  | Tab. 1 | | Tab. 2 | | Harrison and Horner | | Cornwell-Clyne | | Collins and Giles | |
| --- | --- | --- | --- | --- | --- | --- | --- | --- | --- | --- |
|  | Avg | Max | Avg | Max | Avg | Max | Avg | Max | Avg | Max |
| **Tab. 2** | 9.97 | 26.08 | | | | | | | | |
| **Harrison and Horner** | 16.61 | 35.66 | 7.48 | 14.36 | | | | | | |
| **Cornwell-Clyne** | 17.14 | 37.02 | 8.48 | 16.24 | 1.54 | 3.11 | | | | |
| **Collins and Giles** | 15.37 | 33.64 | 6.62 | 12.97 | 1.27 | 2.79 | 2.19 | 3.62 | | |
| **Neblette** | 17.60 | 38.66 | 8.99 | 17.41 | 2.40 | 4.90 | 1.16 | 2.54 | 3.04 | 5.50 |

Tab. 3 – ΔE (CIE 2000) between every pair of reconstructions in Fig. 4.

We compared each image pair using both average and maximum ΔE (CIE2000) values. To mitigate the influence of film damage and other anomalies, we excluded the 1 % of pixels with the largest ΔE values from this comparison. Results are summarized in Tab. 3. ΔE values range from 0 to 100, where ΔE < 1 is barely perceptible; $1 \leq \Delta E < 2$ is visible only upon close inspection; $2 \leq \Delta E < 3$ represent slight (considered acceptable for commercial reproduction) differences; $3 \leq \Delta E < 5$ indicate obvious differences; $\Delta E \geq 5$ easily noticeable differences. The main difference between reconstructions is their overall color cast, caused by imbalances in the primary colors and their relative area in simulated réseau under D65. (Recall that réseau was neutral grey.) We added an option to Color-Screen to automatically adjust the intensities of primaries to obtain that. This correction is shown in Fig. 5 and Tab. 4 where we also tested a second image (Fig. 6) with more warm tones (skin tones and gray grass) to demonstrate the impact of image selection; the first image was chosen for its saturated greens.

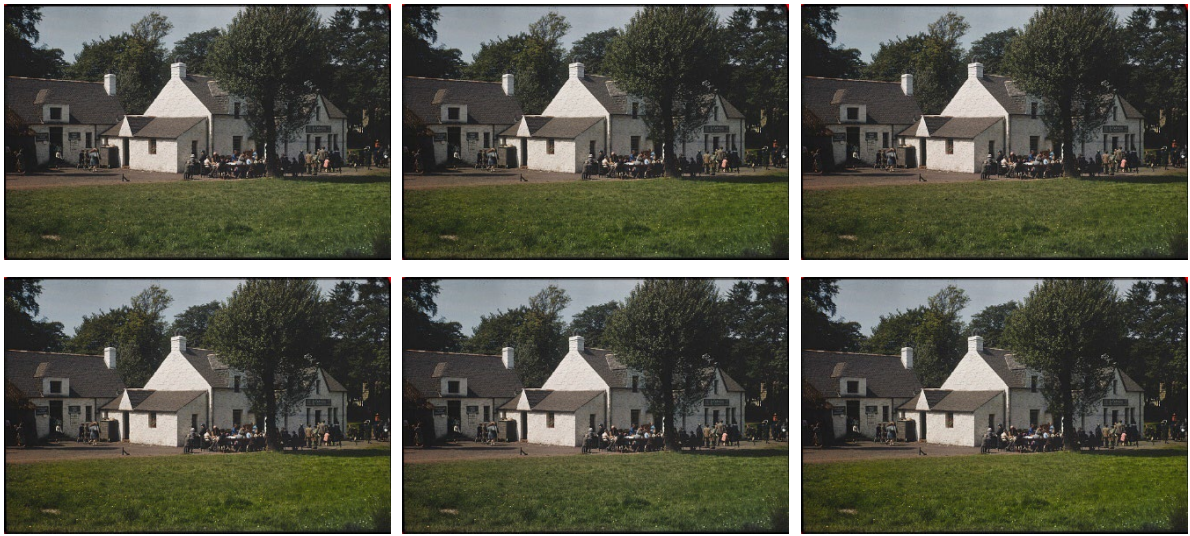

Fig. 5 – Color reconstructions using the same primaries as in Fig. 4 with réseau white-balanced.

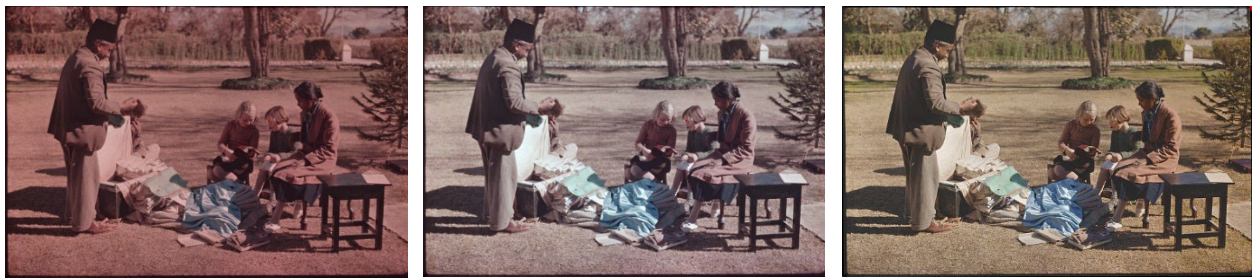

Fig. 6 – Second sample Dufaycolor 6×9 cm approx. 1937, Egypt: (a) color calibrated scan, (b) with color of réseau balanced to neutral using levels function in GIMP, (c) color reconstruction. Likely the same photographer as Fig. 3.

|  |  | Tab. 1 | | Tab. 2 | | Harrison and Horner | | Cornwell-Clyne | | Collins and Giles | |
|---|---|---|---|---|---|---|---|---|---|---|---|
|  |  | Avg | Max | Avg | Max | Avg | Max | Avg | Max | Avg | Max |
| Tab. 2 | Fig. 3 | 0.99 | 2.44 |  |  |  |  |  |  |  |  |
|  | Fig. 6 | 1.46 | 2.87 |  |  |  |  |  |  |  |  |
| Harrison and Horner | Fig. 3 | 0.60 | 1.98 | 0.96 | 2.39 |  |  |  |  |  |  |
|  | Fig. 6 | 0.92 | 2.05 | 1.42 | 2.63 |  |  |  |  |  |  |
| Cornwell-Clyne | Fig. 3 | 1.03 | 2.58 | 0.81 | 3.07 | 1.07 | 3.48 |  |  |  |  |
|  | Fig. 6 | 1.54 | 2.80 | 0.58 | 1.93 | 1.30 | 2.49 |  |  |  |  |
| Collins and Giles | Fig. 3 | 0.76 | 2.65 | 0.81 | 2.45 | 0.38 | 1.12 | 1.16 | 4.18 |  |  |
|  | Fig. 6 | 1.04 | 2.08 | 1.02 | 2.22 | 0.69 | 1.64 | 1.18 | 2.62 |  |  |
| Neblette | Fig. 3 | 2.03 | 4.79 | 1.48 | 4.84 | 2.00 | 5.26 | 1.01 | 2.28 | 2.05 | 5.92 |
|  | Fig. 6 | 2.97 | 5.49 | 1.66 | 4.22 | 2.62 | 4.66 | 1.44 | 2.70 | 2.46 | 4.72 |

Tab. 4 – ΔE (CIE 2000) between every pair of reconstructions in Fig. 5 and reconstructions of image in Fig. 6.

### Conclusions

We compared color renderings based on five different historical measurements. Despite the inherent limitations of colorimetry and spectroscopy from the 1930s to 1950s, which resulted in significant differences between the raw measurements, we found a surprisingly good color match on sample images after white-balancing the réseau: mixing primaries together in real image reduces the differences. The agreement between all pairs of specifications in Harrison and Horner, Cornwell-Clyne, and Collins and Giles yields an average difference less than 1.18 ΔE and a maximum of less than 4.18 ΔE and they match well coloriometric measurements in Tab. 2. This makes the choice of historical measurement less significant than issues related to varying scan sharpness and accurately simulating light passing through the emulsion and réseau—topics we will address in a subsequent paper (with aim to mitigate differences in color reconstructions from scans of same Dufaycolor photograph using different scanners). Notably, even using sRGB primaries resulted in a reasonable average $\Delta E \leq 2.17$ and a maximal $\Delta E \leq 4.83$ in the same test as in Tab. 4, suggesting the possibility of reasonably accurate color reconstructions for other additive color processes. Reconstructions of Dufaycolor based on the infrared scan will probably need some manual white-balancing also due to the fact, that the film was back-coated by a corrective subtractive filter as part of the quality checking. Examining larger set of Dufaycolors will be necessary to evauluate importance of this. We note that even from RGB scans good reconstructions, up to white-balance, are possible using a technique estimating infrared channel using RGB values (Barker *et al.*, 2022). **Acknowledgement**: We thank to Bertrand Lavédrine for discussions that significantly improved this paper.